%% file: main.tex
\documentclass[10pt,twocolumn,letterpaper]{article}

\usepackage[pagenumbers]{cvpr} % To force page numbers, e.g. for an arXiv version

\input{preamble}

\usepackage[accsupp]{axessibility} % Improves PDF readability for those with visual impairments.
\usepackage{multirow}
\usepackage{animate}
\usepackage{color, colortbl}
\definecolor{LightGray}{gray}{0.9}
\usepackage{amsmath}

\DeclareMathOperator*{\argmin}{arg\,min}
\newlength{\itemheight} %

\definecolor{cvprblue}{rgb}{0.21,0.49,0.74}
\usepackage[pagebackref,breaklinks,colorlinks,citecolor=cvprblue]{hyperref}

\def\paperID{2623} % *** Enter the Paper ID here
\def\confName{CVPR}
\def\confYear{2024}

\title{S-DyRF: Reference-Based Stylized Radiance Fields for Dynamic Scenes}

\author{Xingyi Li$^{1,2}$\hspace{0.1in} 
        Zhiguo Cao$^{2}$\hspace{0.1in} 
        Yizheng Wu$^{1,2}$\hspace{0.1in}
        Kewei Wang$^{1,2}$\hspace{0.1in} 
        Ke Xian$^{1,2}$\footnotemark[1]~\hspace{0.1in}
        Zhe Wang$^{3}$\hspace{0.1in} 
        Guosheng Lin$^{1}$\footnotemark[1]\\
$^1$S-Lab, Nanyang Technological University\\
$^2$School of AIA, Huazhong University of Science and Technology\hspace{0.2in}$^3$SenseTime Research\\
{\tt\small \{xingyi\_li,zgcao,kxian\}@hust.edu.cn, gslin@ntu.edu.sg}\\
{\small{\url{https://xingyi-li.github.io/s-dyrf}}}
\vspace{-2mm}
}

\begin{document}
% \maketitle

\newcommand{\doanimated}

\twocolumn[{%
\renewcommand\twocolumn[1][]{#1}%
\maketitle
\centering
\setlength{\itemheight}{3.35cm}
\ifdefined\doanimated

\animategraphics[autoplay,loop,trim = 0 0 0 0,height=\itemheight]{20}{figures/teaser/sear_steak_pencil_v2/}{1}{120} 
\animategraphics[autoplay,loop,trim = 0 0 0 0,height=\itemheight]{20}{figures/teaser/cook_spinach/}{1}{120} 
\animategraphics[autoplay,loop,trim = 0 0 0 0,height=\itemheight]{20}{figures/teaser/cut_roasted_beef/}{1}{120} 
\animategraphics[autoplay,loop,height=\itemheight]{20}{figures/teaser/mutant/}{1}{120} 
\else
    \includegraphics[trim = 0 0 0 0,height=\itemheight]{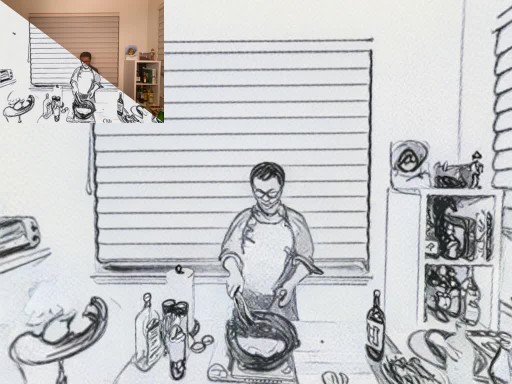}  
    \includegraphics[clip,trim = 0 0 0 0,height=\itemheight]{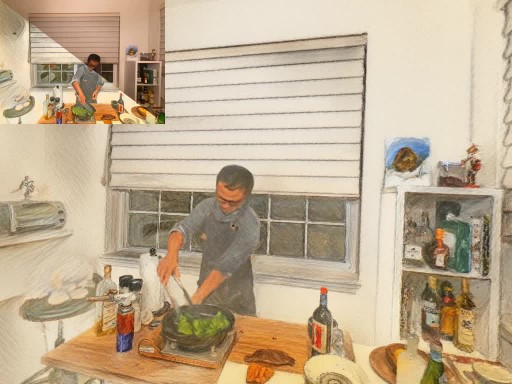} 
    \includegraphics[clip,trim = 0 0 0 0,height=\itemheight]{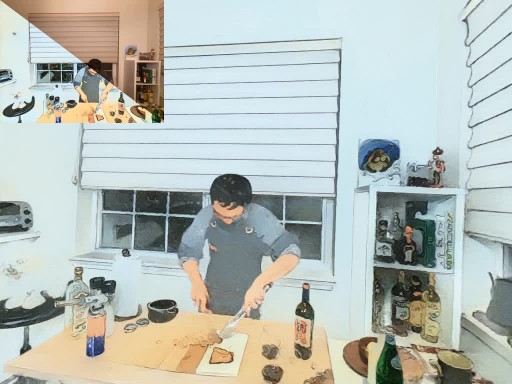}
    \includegraphics[clip,height=\itemheight]{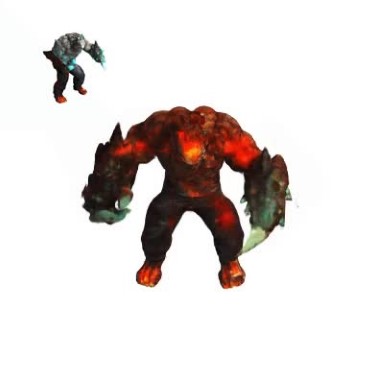} 
\fi
% \vspace{-0.5cm}
\small{\captionof{figure}{\label{fig:teaser} 
Our method can stylize dynamic 3D scenes, maintaining semantic consistency with the given reference image across both spatial and temporal dimensions. Besides novel time synthesis (the leftmost one), our stylized dynamic radiance field can synthesize novel views (the second one) or perform space-time view synthesis (the third one)~\cite{li20233d,shen2023make,li2021neural}. Furthermore, our method can also stylize synthetic objects (the rightmost one). We encourage readers to experience the animations by viewing them with Adobe Acrobat or KDE Okular.
}}
\vspace{1.2em}
% \vspace{2em}
}]

\renewcommand{\thefootnote}{\fnsymbol{footnote}}
\footnotetext[1]{Corresponding author.}

\input{sec/0_abstract}    
\input{sec/1_intro}
\input{sec/2_relatedwork}

\input{sec/4_method}
\input{sec/5_exp}

\input{sec/6_conclusion}

\noindent\textbf{Acknowledgements.}
This study is supported under the RIE2020 Industry Alignment Fund – Industry Collaboration Projects (IAF-ICP) Funding Initiative, as well as cash and in-kind contribution from the industry partner(s). This work is also supported by the MOE AcRF Tier 2 grant (MOE-T2EP20220-0007).

{
    \small
    \bibliographystyle{ieeenat_fullname}
    \bibliography{main}
}


\end{document}

%% file: preamble.tex
\usepackage[dvipsnames]{xcolor}

%% file: sec/0_abstract.tex
\begin{abstract}
    Current 3D stylization methods often assume static scenes, which violates the dynamic nature of our real world. To address this limitation, we present S-DyRF, a reference-based spatio-temporal stylization method for dynamic neural radiance fields. However, stylizing dynamic 3D scenes is inherently challenging due to the limited availability of stylized reference images along the temporal axis. Our key insight lies in introducing additional temporal cues besides the provided reference. To this end, we generate temporal pseudo-references from the given stylized reference. These pseudo-references facilitate the propagation of style information from the reference to the entire dynamic 3D scene. For coarse style transfer, we enforce novel views and times to mimic the style details present in pseudo-references at the feature level. To preserve high-frequency details, we create a collection of stylized temporal pseudo-rays from temporal pseudo-references. These pseudo-rays serve as detailed and explicit stylization guidance for achieving fine style transfer. 
    Experiments on both synthetic and real-world datasets demonstrate that our method yields plausible stylized results of space-time view synthesis on dynamic 3D scenes.
\end{abstract}

%% file: sec/1_intro.tex
\vspace{-1.5em}
\section{Introduction}
\label{sec:intro}

Style transfer~\cite{gatys2016image,johnson2016perceptual,kolkin2022neural,huang2017arbitrary,li2017universal,svoboda2020two,li2019learning} involves the process of reimagining an image's content by emulating the artistic style of another. While it can produce high-quality stylized images, it confines the results to an identical viewpoint as the content image. In recent years, there has been a burgeoning demand for the stylization and editing of 3D scenes and objects across a range of industries, encompassing the realms of gaming, cinematic experiences, and mixed reality applications. Previous methods primarily concentrate on stylizing 3D objects and scenes using point clouds~\cite{huang2021learning,cao2020psnet,segu20203dsnet,mu20223d}, meshes~\cite{kato2018neural,hollein2022stylemesh}, or applying style transfer to both the geometry and texture~\cite{yin20213dstylenet}.

As neural radiance fields continue to evolve at a rapid pace~\cite{mildenhall2020nerf,chen2022tensorf,muller2022instant,cao2023hexplane,li2022symmnerf}, recent advancements~\cite{huang2021learning,huang2022stylizednerf,wang2022clip,wang2023nerf,fan2022unified,nguyen2022snerf,chen2022upst,chiang2022stylizing,liu2023stylerf,haque2023instruct,shao2023control4d} have incorporated neural radiance fields into 3D stylization, significantly easing the process of transferring style from any 2D style reference to 3D scenes. While these methods excel at producing high-quality geometrically consistent 3D content, they often fall short in providing users, including artists and designers, with fine-grained control over the stylization process, which can be limiting in achieving the precise aesthetic effects desired. To address this challenge, several methods have emerged as promising solutions. For example, Ref-NPR~\cite{zhang2023ref} introduces a reference-guided, controllable approach to scene stylization. This method empowers users to create stylized views while ensuring both geometric and semantic consistency with a provided stylized reference image. Despite a significant step towards enhancing the controllability of stylization in 3D scenes, these methods typically rely on the assumption that the scene is static. However, in our dynamic world, scenes often feature diverse dynamic content, underscoring the increasing need to develop stylization methods suitable for dynamic 3D scenes.

In this paper, our objective is to take the first step towards reference-based stylized neural radiance fields for dynamic 3D scenes. The challenge at hand involves not only stylizing the 3D scenes but also addressing the temporal dimension, a factor that distinguishes it from the stylization of static 3D scenes. One intuitive approach to address this challenge is directly applying video-based stylization methods~\cite{jamrivska2019stylizing,texler2020interactive,chen2017coherent,huang2017real}. While these methods excel at generating stylized videos that faithfully preserve the texture of the original style exemplar, they may deviate from the desired style when applied to novel view synthesis and can suffer from flickering issues.

To tackle the challenges outlined above, we introduce a novel paradigm $\textit{S-DyRF}$ for stylizing dynamic 3D scenes, utilizing either a single or a small number of stylized reference images. Considering the scarcity of stylized reference images along the time axis, our key insight lies in the introduction of additional temporal cues to propagate style information throughout different timestamps. To achieve this, our method starts by generating temporal pseudo-references that act as effective guidance to ensure style consistency across the temporal domain. Next, we establish a mapping of content and style, followed by transferring the style information contained within those pseudo-references to novel views and times at the feature level. To preserve details, we also draw inspiration from Ref-NPR~\cite{zhang2023ref} and introduce the Temporal Reference Ray Registration process. Through this process, a collection of stylized temporal pseudo-rays is generated from the pseudo-references. These pseudo-rays then act as a powerful tool for offering detailed and explicit stylization guidance for achieving fine style transfer. Bringing everything together, our S-DyRF is capable of synthesizing stylized novel views and times that uphold semantic consistency with the given reference image across both spatial and temporal dimensions. 

In summary, our main contributions are:
\begin{itemize}
    \item We propose a new task of stylizing dynamic 3D scenes using one or a small number of stylized reference images.
    \item We introduce S-DyRF, a reference-based spatio-temporal stylization method for dynamic neural radiance fields.
    \item Our method showcases superior qualitative and quantitative results, as well as providing controllability for customizing dynamic 3D scenes.
\end{itemize}

%% file: sec/2_relatedwork.tex
\section{Related Work}
\label{sec:related-work}

\begin{figure*}[htbp]
    \centering
    \includegraphics[width=0.9\textwidth]{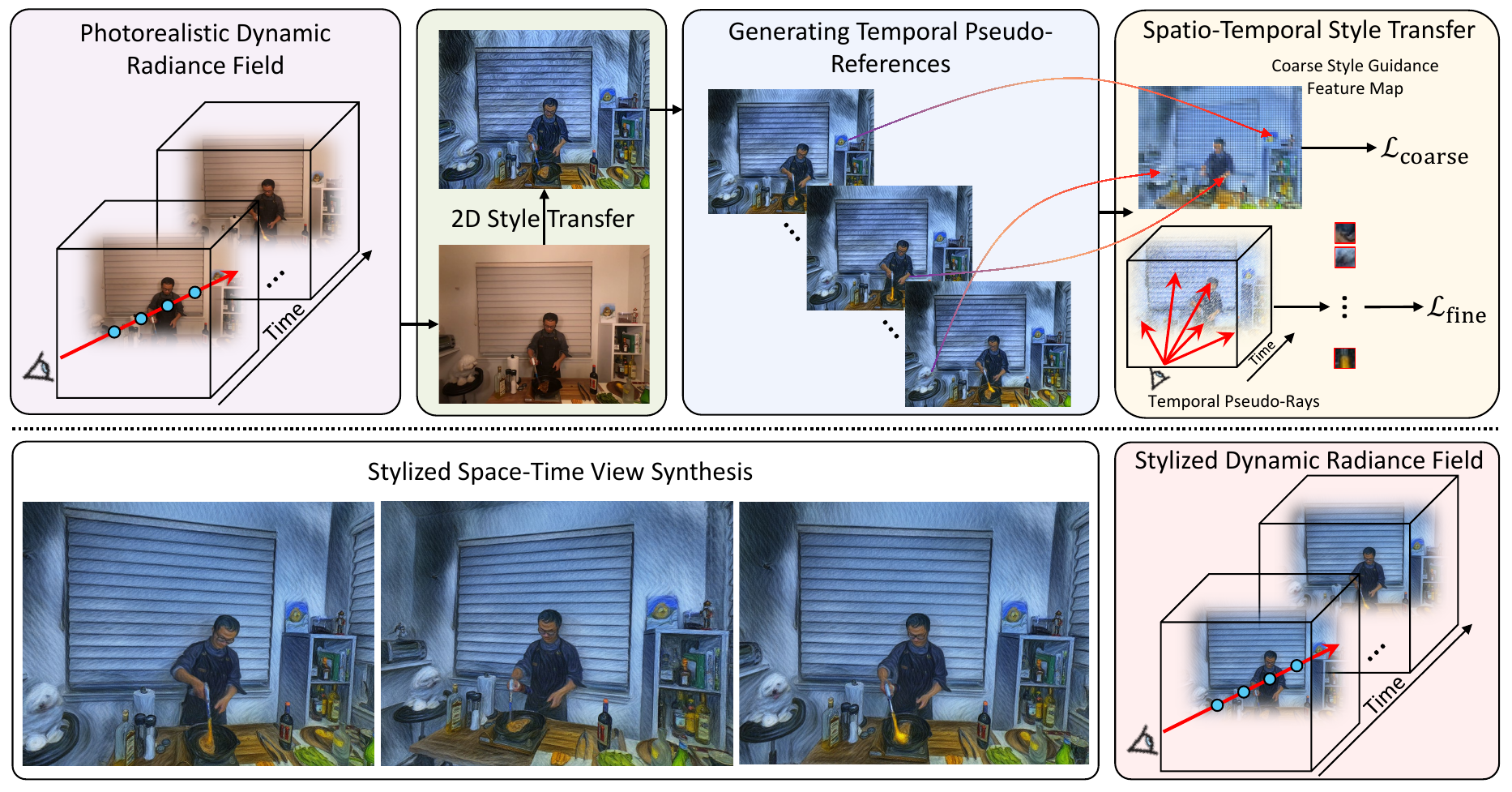}
    \caption{\textbf{An overview of our method.} Given a pre-trained photorealistic dynamic radiance field, we first render a reference view at time $k$ from a specific reference camera. Following that, the reference view undergoes a 2D style transfer using an appropriate method, e.g., manual editing, NNST~\cite{kolkin2022neural}, or ControlNet~\cite{zhang2023adding}, to produce a stylized reference image. To propagate the style information from the stylized reference to other timestamps, we generate temporal pseudo-references and apply spatio-temporal style transfer to optimize our dynamic radiance field. Once this stylization is done, we can yield plausible stylized results of space-time view synthesis on dynamic 3D scenes.}
    \vspace{-2mm}
    \label{fig:pipeline}
\end{figure*}

\noindent\textbf{Image and video style transfer.}
Style transfer involves the process of recomposing the content of an image to emulate the style of another. Pioneering endeavors in this field can be traced back to image analogies~\cite{hertzmann2001image}. This approach and follow-up work~\cite{liao2017visual,he2019progressive,fivser2016stylit} deviate from manual filter programming and, instead, automatically learn filters from examples. However, they necessitate a certain level of semantic similarity between the content and style images. Since Gatys et al.~\cite{gatys2016image} 
initially proposed the idea of utilizing high-level features extracted from pre-trained deep neural networks to represent image style, 
significant progress ~\cite{johnson2016perceptual,kolkin2022neural,huang2017arbitrary,li2017universal,svoboda2020two,li2019learning} has been made in the field of neural style transfer. In contrast to example-based methods, neural style transfer exhibits the capability to apply artistic stylization to various images, offering greater versatility in the stylization process. Although exhibiting similarities with image style transfer, video style transfer methods~\cite{jamrivska2019stylizing,texler2020interactive,chen2017coherent,huang2017real} primarily concentrate on maintaining temporal consistency throughout the entire video sequence. As a representative method of video style transfer, Texler et al.~\cite{texler2020interactive} develop an appearance translation network from the beginning, utilizing just a handful of stylized exemplars to extend the desired style across the entire video sequence. Nonetheless, these approaches primarily target enhancing short-term consistency between adjacent frames, and they lack the capability to facilitate novel view synthesis.

\noindent\textbf{3D style transfer.}
Despite the extensive exploration of style transfer within the image domain, the field of 3D style transfer is still in its infancy. Prior methods have mainly focused on stylizing individual 3D objects, utilizing point clouds~\cite{segu20203dsnet}, meshes~\cite{kato2018neural}, or implementing style transfer on both geometry and texture~\cite{yin20213dstylenet}. Recent developments have also directed attention towards stylizing entire 3D scenes. This can be achieved by explicitly applying style transfer on point cloud~\cite{huang2021learning,cao2020psnet,mu20223d} or mesh~\cite{hollein2022stylemesh} representations. However, these methods may encounter issues arising from imperfect geometry and texture rendering and are limited to static scenes.

\noindent\textbf{Stylizing radiance fields.}
Fueled by the remarkable advancements in novel view synthesis achieved by neural radiance fields~\cite{mildenhall2020nerf,wu2022dof}, the research community has been energized to delve deeper into the emerging domain of stylizing radiance fields~\cite{huang2021learning,huang2022stylizednerf,wang2022clip,wang2023nerf,fan2022unified,nguyen2022snerf,chen2022upst,chiang2022stylizing,liu2023stylerf,shao2023control4d,haque2023instruct}, i.e., transferring style to 3D scenes represented as neural radiance fields. In particular, ARF~\cite{zhang2022arf} introduces a nearest neighbor-based loss that can capture style details while maintaining multi-view consistency. Although ARF excels in producing high-quality artistic 3D content, it does not afford explicit control over the stylized results. To control the stylization, Pang et al.~\cite{pang2023locally} propose employing hash-grid encoding for learning the appearance and geometry embeddings, thereby facilitating local style transfer. For enhanced user control over the stylization process, Ref-NPR~\cite{zhang2023ref} introduces a reference-guided, controllable scene stylization method. This method empowers users to generate stylized novel views that maintain geometric and semantic consistency with a given stylized reference image. However, these methods are limited to stylizing static scenes. Therefore, we are interested in taking the first step towards reference-based stylized neural radiance fields for dynamic 3D scenes.

%% file: sec/4_method.tex
\section{Method}
\label{sec:method}

\subsection{Overview}
We aim to stylize dynamic 3D scenes using one or a few stylized reference images as input. The expected output of our method is stylized novel views and times that maintain semantic consistency with the given reference image across both spatial and temporal dimensions. Fig.~\ref{fig:pipeline} presents a schematic depiction of our pipeline. We assume a pre-trained photorealistic dynamic radiance field $F_{\Theta}$ (e.g., HexPlane~\cite{cao2023hexplane}) that is reconstructed from multiple images or videos capturing complex, real-world, dynamic 3D scenes, and denote our stylized dynamic radiance field as $G_{\Theta}$. To simplify our discussion without sacrificing generality, we focus on scenarios involving only a single reference image. We first render a reference view $\mathcal{I}_{F_{\Theta},R}^{k}$ at time $k$ from a specific reference camera $\mathbf{p}_{R}$. Following that, the reference view $\mathcal{I}_{F_{\Theta},R}^{k}$ undergoes a 2D stylization process using an appropriate method, e.g., manual editing, NNST~\cite{kolkin2022neural}, or ControlNet~\cite{zhang2023adding}, to produce a stylized reference image $\mathcal{S}_{R}^{k}$. 

Given the limited number of stylized reference images along the time axis, we propose to introduce additional temporal cues and explicitly propagate the style information from the stylized reference to other timestamps to enable a more coherent and consistent stylization across the entire time series. To achieve this, our method starts by generating temporal pseudo-references that serve as effective sources of supervision for maintaining style consistency across the temporal domain. We then establish a coarse content-style mapping and broadcast the style information contained in pseudo-references to novel views and times at the feature level, facilitating a coarse style transfer. To preserve high-frequency details, we also draw inspiration from Ref-NPR~\cite{zhang2023ref} and introduce the Temporal Reference Ray Registration process. This process produces a set of stylized temporal pseudo-rays that are generated from pseudo-references. These pseudo-rays serve as a means to provide explicit stylization supervision for fine style transfer.

\subsection{Generating Temporal Pseudo-References}
We focus on stylizing dynamic 3D scenes by leveraging stylized reference images, allowing for greater controllability. However, the challenge arises when dealing with a limited number of stylized reference images, particularly in the context of stylizing dynamic radiance fields, mainly because there is a lack of explicit supervision for the majority of timestamps. For consistent stylization over time, our key insight involves the introduction of additional temporal cues and the explicit propagation of style information from the stylized reference to other timestamps. 

To achieve this, we leverage a video style transfer method to generate temporal pseudo-references. The process begins by rendering novel times, denoted as $\{\mathcal{I}_{F_{\Theta},R}^{i}\}_{i=1}^{T}$ that share an identical camera pose $\mathbf{p}_{R}$ with the stylized reference image $\mathcal{S}_{R}^{k}$ using the pre-trained dynamic radiance field $F_{\Theta}$. To train, we randomly sample small patches from $\mathcal{I}_{F_{\Theta},R}^{k}$, pass them through a U-net-based image-to-image translation framework~\cite{futschik2019real}, and generate their stylized versions. Subsequently, we calculate the loss by comparing these stylized patches with the corresponding patches extracted from the stylized reference. This error is then backpropagated through the network to optimize the model parameters. Once the model is trained, we can feed those novel times $\{\mathcal{I}_{F_{\Theta},R}^{i}\}_{i=1}^{T}$ into the network and synthesize temporal pseudo-references $\{\mathcal{\hat{S}}_{R}^{i}\}_{i=1}^{T}$. We end up replacing $\mathcal{\hat{S}}_{R}^{k}$ in $\{\mathcal{\hat{S}}_{R}^{i}\}_{i=1}^{T}$ with $\mathcal{S}_{R}^{k}$. These pseudo-references serve as effective sources of supervision, ensuring consistent stylization across the temporal domain, thereby aiding in the optimization of the dynamic radiance field.

\subsection{Spatio-Temporal Style Transfer}
We now have temporal pseudo-references besides the provided stylized reference. Our next step involves performing spatio-temporal style transfer that leverages these temporal pseudo-references to propagate style information from the stylized reference across both temporal and spatial dimensions. This spatio-temporal style transfer encompasses coarse and fine style transfer.

\noindent\textbf{Coarse style transfer.}
We first transfer style information over time and space by ensuring that the semantic correspondence across the entire dynamic scene remains consistent throughout the stylization process. To accomplish this, we establish a coarse content-style mapping and enforce novel views and times to mimic the style details contained in pseudo-references at the feature level. Given the inherent redundancy of style information in temporal pseudo-references, we select only $N$ pseudo-references from $\{\mathcal{\hat{S}}_{R}^{i}\}_{i=1}^{T}$ for coarse style transfer, denoted as $\{\mathcal{\hat{S}}_{R}^{j}\}_{j=1}^{N}$. We first render a content domain image $\mathcal{I}^{t}_{F_{\Theta}}$ from the pre-trained dynamic radiance field $F_{\Theta}$, given a time $t$ and camera pose $\mathbf{p}$. Subsequently, we input $\mathcal{I}^{t}_{F_{\Theta}}$, $\{\mathcal{\hat{S}}_{R}^{j}\}_{j=1}^{N}$ and their corresponding content domain images $\{\mathcal{I}_{F_{\Theta},R}^{j}\}_{j=1}^{N}$ into a pre-trained feature extractor, yielding high-level semantic features $\mathcal{F}_{\mathcal{I}^{t}_{F_{\Theta}}}$, $\{\mathcal{F}_{\mathcal{\hat{S}}^{j}_R}\}_{j=1}^{N}$ and $\{\mathcal{F}_{\mathcal{I}_{F_{\Theta},R}^{j}}\}_{j=1}^{N}$. To transfer style information both temporally and spatially, we construct a coarse style guidance feature map $\mathcal{F}_{\mathcal{G}^{t}}$ from $\{\mathcal{F}_{\mathcal{\hat{S}}^{j}_R}\}_{j=1}^{N}$. Specifically, we compute the feature vector at $(m, n)$ of $\mathcal{F}_{\mathcal{G}^{t}}$ as follows:
\begin{gather}
    \mathcal{F}_{\mathcal{G}^{t}}^{\left(m, n\right)} = \oplus \left( \{\mathcal{F}_{\mathcal{\hat{S}}^{j}_R}\}_{j=1}^{N} \right)^{\left(p^{*}, q^{*}\right)},\\
    \left(p^{*}, q^{*}\right) = \argmin_{p, q} d \left( \mathcal{F}_{\mathcal{I}^{t}_{F_{\Theta}}}^{(m, n)}, \oplus \left( \{\mathcal{F}_{\mathcal{I}_{F_{\Theta},R}^{j}}\}_{j=1}^{N} \right)^{\left(p, q\right)}  \right),
\end{gather}
where the concatenation operation $\oplus$ is performed along the width dimension, $d$ is the cosine distance. The resulting style guidance feature map c contains spatio-temporal style information, which can be employed to optimize the dynamic radiance field, propagating style information from the stylized reference across both temporal and spatial dimensions. To achieve this, we minimize the cosine distance $d$ between $\mathcal{F}_{\mathcal{G}^{t}}$ and the extracted feature map $\mathcal{F}_{\mathcal{S}^{t}_{G_{\Theta}}}$ of rendered view $\mathcal{S}^{t}_{G_{\Theta}}$ from our stylized dynamic radiance field $G_{\Theta}$, given a time $t$ and camera pose $\mathbf{p}$. Following prior works~\cite{zhang2023ref,zhang2022arf}, we also incorporate a feature-level content-preserving loss to ensure that the content remains easily recognizable. The feature-level stylization loss is defined as:
\begin{equation}
    \mathcal{L}_{\text{feat}} = \sum_{i,j} \left( d\left(\mathcal{F}_{\mathcal{G}^{t}}, \mathcal{F}_{\mathcal{S}^{t}_{G_{\Theta}}}\right) + \lambda \Big\Vert  \mathcal{F}_{\mathcal{I}^{t}_{F_{\Theta}}} - \mathcal{F}_{\mathcal{S}^{t}_{G_{\Theta}}} \Big\Vert^{2}_{2} \right).
\end{equation}

We also add a coarse color-level stylization loss to further encourage color-matching. Specifically, we first downsample each of $\{\mathcal{\hat{S}}_{R}^{j}\}_{j=1}^{N}$ to the size of $\mathcal{F}_{\mathcal{G}^{t}}$ and concatenate them together, denoted as $\oplus \{\mathcal{\hat{S}}_{R, D}^{j}\}_{j=1}^{N}$, and use the aforementioned content-style mapping to obtain a coarse style guidance color map $\mathcal{S}_{\mathcal{G},D}^{t}$. We also downsample $\mathcal{S}^{t}_{G_{\Theta}}$, denoted as $\mathcal{S}^{t}_{G_{\Theta},D}$. The coarse color-level stylization loss is then computed as follows:
\begin{equation}
    \mathcal{L}_{\text{color}} = \sum_{i,j} \Big\Vert \mathcal{S}_{\mathcal{G},D}^{t} - \mathcal{S}^{t}_{G_{\Theta},D} \Big\Vert^{2}_{2}.
\end{equation}
Bringing everything together, our coarse style transfer loss can be expressed as follows:
\begin{equation}
    \mathcal{L}_{\text{coarse}} = \lambda_{\text{feat}}\mathcal{L}_{\text{feat}} + \lambda_{\text{color}}\mathcal{L}_{\text{color}}.
\end{equation}
As suggested in ARF~\cite{zhang2022arf}, rendering a full-resolution image at each optimization iteration is impractical due to its memory inefficiency. To tackle this issue, we adopt the deferred back-propagation technique employed in ARF~\cite{zhang2022arf} to optimize the coarse style transfer loss $\mathcal{L}_{\text{coarse}}$.

\noindent\textbf{Fine style transfer.}
While coarse style transfer already yields a fairly good stylization result, it lacks high-frequency details. To retain high-frequency details, we draw inspiration from Ref-NPR~\cite{zhang2023ref} and introduce the Temporal Reference Ray Registration process. Similar to Ref-NPR, we utilize pseudo-depth rendered from the pre-trained dynamic radiance field to obtain reference-related explicit supervision. However, since we deal with dynamic 3D scenes, we need to account for the temporal dimension as well. This can be achieved by introducing temporal pseudo-references. Specifically, given time $t$ and reference camera pose $\mathbf{p}_{R}$, we first lift the pseudo-reference $\mathcal{\hat{S}}_{R}^{t}$ into 3D space using corresponding depth values rendered from the pre-trained dynamic radiance field, and maps these 3D positions to corresponding voxels. Formally, we define the temporal reference dictionary $D$ as
\begin{gather}
    D(x,y,z,t) = \{  \mathbf{r}_{i} \in \mathcal{R} \, | \, Q(\mathbf{x}(\mathbf{r}_{i}, t)) = (x, y, z) \},
\end{gather}
where $\mathcal{R}$ is the set of rays originating from the reference camera pose $\mathbf{p}_{R}$, $\mathbf{x}(\mathbf{r}_{i}, t)$ denotes the intersection point of the ray $\mathbf{r}_{i}$ with the surface at time $t$, and $Q$ is a quantization operator responsible for mapping 3D positions $\mathbf{x}(\mathbf{r}_{i}, t)$ to their corresponding voxels. Given time $t$, the notation $D(x,y,z,t)$ represents the collection of rays that drop into the voxel located at $(x, y, z)$ and the color of $D(x,y,z,t)$ is given by pseudo-references $\mathcal{\hat{S}}_{R}^{t}$. Following this, our temporal reference ray registration process is to find those rays $\mathbf{r}_{j} \in \mathcal{T}$ emitted from training views a pseudo-ray $\hat{\mathbf{r}}_{j} \in \mathcal{R}$. $\mathbf{r}_{j}$ should have an intersection point $\mathbf{x}(\mathbf{r}_{j}, t)$ falling into its nearest voxel $D(x,y,z,t)$ and share a similar direction $\mathbf{d}_{\mathbf{r}_{j}}$ with that of $\mathbf{r}_{i}$ contained in $D(x,y,z,t)$. These pseudo-rays are then assigned with stylized colors $\hat{C}_{R}(\hat{\mathbf{r}}_{j}, t)$ from $D(x,y,z,t)$. The resulting set of temporal pseudo-rays and their corresponding stylized colors are defined as:
\begin{gather}
    \Phi_{R}(t) = \{  (\mathbf{r}_{j}, \hat{C}_{R}(\hat{\mathbf{r}}_{j}, t)) \, | \, \mathbf{r}_{j} \in \mathcal{T} \cup \mathcal{R}, \hat{\mathbf{r}}_{j} \ne \varnothing \},\\
    \hat{\mathbf{r}}_{j} = \argmin_{\mathbf{r}_{i} \in D(x,y,z,t)} \Big\Vert \mathbf{x}(\mathbf{r}_{i}, t) - \mathbf{x}(\mathbf{r}_{j}, t)  \Big\Vert_{2},\\
    \text{s.t.}\quad \angle (\mathbf{d}_{\mathbf{r}_{i}},\mathbf{d}_{\mathbf{r}_{j}}) < \theta, Q(\mathbf{x}(\mathbf{r}_{j}, t)) = (x, y, z).
\end{gather}
To maintain details, we minimize the mean squared error between the rendered color $\hat{C}_{G_{\Theta}}(\mathbf{r}_{i}, t)$ of our model and the color of the corresponding temporal pseudo-ray $\hat{C}_{R}(\hat{\mathbf{r}}_{i}, t)$, which is given by
\begin{equation}
    \mathcal{L}_{\text{fine}} = \sum_{\mathbf{r}_{i} \in \phi_{R}(t)} \Big\Vert  \hat{C}_{G_{\Theta}}(\mathbf{r}_{i}, t) - \hat{C}_{R}(\hat{\mathbf{r}}_{i}, t) \Big\Vert_{2}^{2},
\end{equation}
where $\phi_{R}(t)$ represents the set of those temporal pseudo-rays at time $t$.

\subsection{Optimization}
This section describes our training scheme. We adopt a hierarch stylization strategy that gradually propagates the style information from the stylized reference to the entire dynamic 3D scene. Specifically, we employ hierarchical stylization on keyframes and subsequently apply it to the entire sequence.

For a 4D point represented as $(x, y, z, t)$, we calculate its volume density and appearance features using a dynamic radiance field. Subsequently, we predict the final color by feeding the appearance feature as inputs to an MLP. We use standard volume rendering to accumulate colors and densities into 2D images. To stylize dynamic 3D scenes, our total optimization objective is defined as:
\begin{gather}
    \mathcal{L} = \lambda_{\text{c}}\mathcal{L}_{\text{coarse}} + \lambda_{\text{f}}\mathcal{L}_{\text{fine}} + \lambda_{\text{TV}}\mathcal{L}_{\text{TV}},
\end{gather}
where $\mathcal{L}_{\text{TV}}$ is the Total Variational (TV) loss that serves to enforce the spatio-temporal continuity.

%% file: sec/5_exp.tex
\begin{figure*}[htbp]
    \centering
    \includegraphics[width=0.91\textwidth]{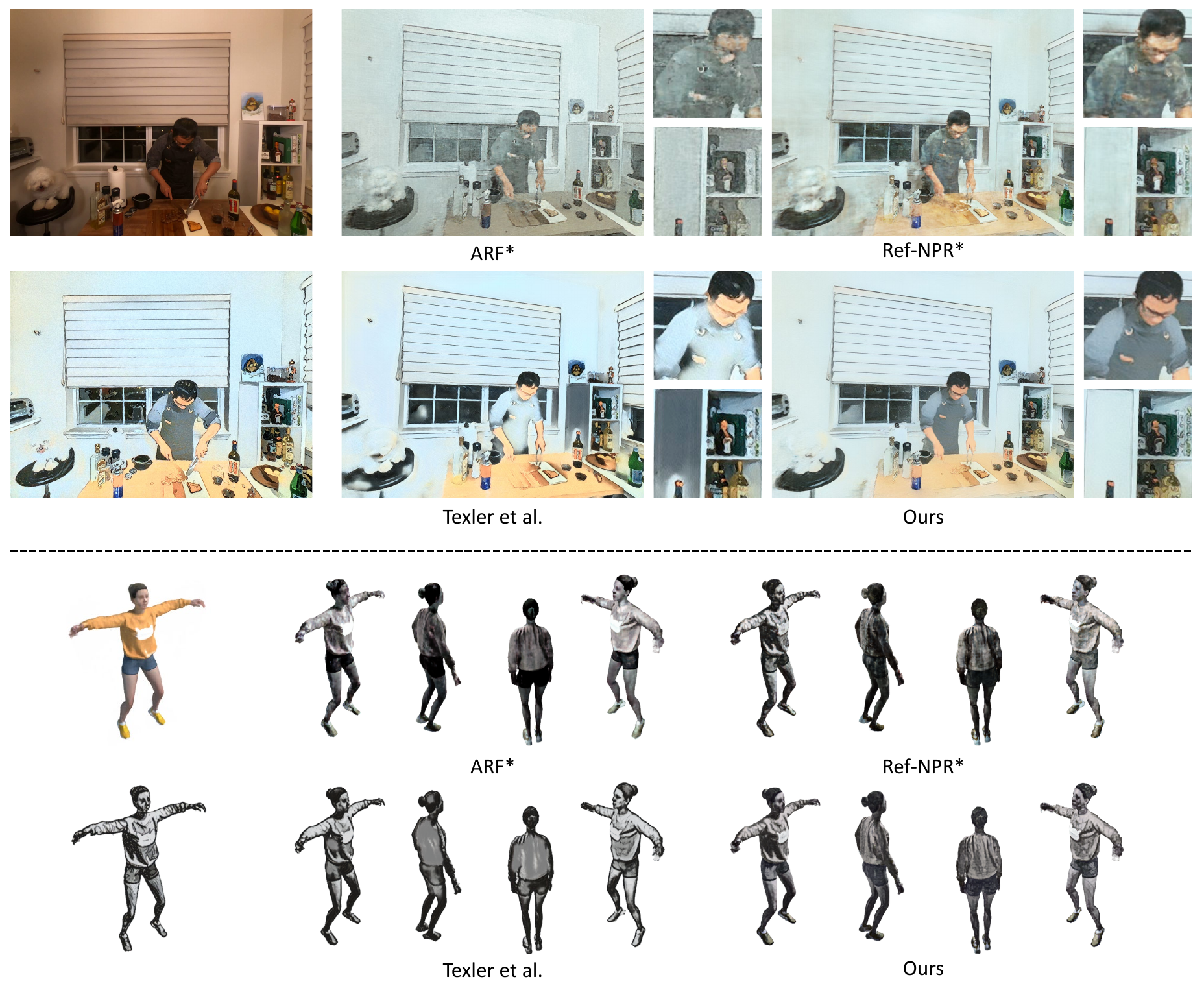}
    \caption{\textbf{Qualitative comparisons on real-world and synthetic datasets.} We compare our method with ARF*~\cite{zhang2022arf}, Ref-NPR*~\cite{zhang2023ref}, and Texler et al.~\cite{texler2020interactive}. In each case, the upper left image represents the reference view, generated from the photorealistic dynamic radiance field, while the lower left image depicts its corresponding stylized version. 
    % When compared to comparative alternatives, our method excels at generating visually pleasing stylized novel views and times while preserving semantic consistency with the provided reference image across both spatial and temporal dimensions.
    }
    \vspace{-2mm}
    \label{fig:qualitative}
\end{figure*}

\section{Experiments}
\label{sec:exp}

\subsection{Implementation Details}
Our dynamic radiance field is based on HexPlane~\cite{cao2023hexplane}, an explicit representation designed for modeling dynamic 3D scenes. The choice of HexPlane stems from its simplicity and effectiveness in representing 4D volumes.
% It is worth noting that our method is versatile and not tied to a specific radiance field representation. 
Following prior works~\cite{zhang2022arf,zhang2023ref}, in the stylization process, we keep the density component of the dynamic radiance field frozen and exclusively optimize the appearance component. To avoid view-dependent stylization, we set and fix the view directions to zero when passing them into the MLP and proceed to finetune the model on training views for a few iterations before stylization.

We leverage a video style transfer method~\cite{texler2020interactive} to generate temporal pseudo-references. For the feature extraction, we adopt pre-trained VGG16~\cite{simonyan2014very} as our feature extractor. As in Ref-NPR~\cite{zhang2023ref}, we set the spatial resolution of the temporal reference dictionary $D$ to $256^3$, where each voxel can at most store $8$ rays. We train our model using the Adam optimizer~\cite{kingma2014adam} and carry out all experiments on a single NVIDIA GeForce RTX 4090 GPU. For hyperparameters, we set $\lambda = 5 \times 10^{-3}$, $\lambda_{\text{TV}} = 5 \times 10^{-2}$, $\lambda_{\text{feat}} = \lambda_{\text{c}} = 1$, $\lambda_{\text{color}} = 5$, and $\lambda_{\text{f}} = 10$.

\subsection{Baselines}
To the best of our knowledge, we are the first to tackle the task of stylizing neural radiance fields for dynamic 3D scenes with a stylized 2D view as a reference. We compare our method with three representative stylization methods. One is ARF~\cite{zhang2022arf}, an arbitrary style transfer method for static 3D scenes. To ensure a fair comparison, we reimplement ARF and make specific adjustments to cater to dynamic scenes. This modified version is denoted as ARF*. Ref-NPR~\cite{zhang2023ref} is a reference-based 3D stylization method with the ability to produce stylized novel views while preserving both geometric and semantic consistency given a stylized reference. Similarly, since it was initially designed for static 3D scenes, we also reimplement it and adapt it to dynamic scenes, which we denote as Ref-NPR*. Finally, we also include Texler et al.~\cite{texler2020interactive}, a reference-based
video stylization method that is capable of extending the stylized content from a stylized reference to the entire video sequence.

\subsection{Results}

\begin{table*}
    \footnotesize
    \centering 
    \caption{{\bf Quantitative comparisons on the Plenoptic Video dataset.} The best performance is in \textbf{bold}, and the second best is \underline{underlined}.}
    \resizebox{0.8\linewidth}{!}{
        \renewcommand\arraystretch{0.85}
		\begin{tabular}{lcccccc}
            \toprule
            \multirow{3}{*}{Method} & \multicolumn{3}{c}{Fixing camera viewpoint} & \multicolumn{3}{c}{Fixing time}\\
            \cmidrule(lr){2-4} \cmidrule(lr){5-7}
            & \multirow{2}{*}{Ref-LPIPS$\downarrow$} & Short-range & Long-range & \multirow{2}{*}{Ref-LPIPS$\downarrow$} & Short-range & Long-range\\
            & & consistency$\downarrow$ & consistency$\downarrow$ & & consistency$\downarrow$ & consistency$\downarrow$ \\
            \midrule
            ARF*~\cite{zhang2022arf} & 0.513 & \textbf{0.006} & \textbf{0.007} & 0.657 & \textbf{0.015} & \textbf{0.072}\\
            Ref-NPR*~\cite{zhang2023ref} & 0.374 & \underline{0.012} & \underline{0.013} & 0.602 & 0.020 & 0.097\\
            Texler et al.~\cite{texler2020interactive} & \underline{0.353} & 0.015 & 0.016 & \underline{0.585} & 0.027 & 0.132\\
            \rowcolor{LightGray}
            Ours & \textbf{0.298} & \underline{0.012} & \underline{0.013} & \textbf{0.561} & \underline{0.018} & \underline{0.084}\\
            \bottomrule
		\end{tabular}
    }
    \vspace{-2mm}
    \label{tab:nv3d-quantitative}
\end{table*}

\begin{table*}
    \footnotesize
    \centering 
    \caption{{\bf Quantitative comparisons on the D-NeRF dataset.} The best performance is in \textbf{bold}, and the second best is \underline{underlined}.}
    \resizebox{0.8\linewidth}{!}{
        \renewcommand\arraystretch{0.85}
		\begin{tabular}{lcccccc}
            \toprule
            \multirow{3}{*}{Method} & \multicolumn{3}{c}{Fixing camera viewpoint} & \multicolumn{3}{c}{Fixing time}\\
            \cmidrule(lr){2-4} \cmidrule(lr){5-7}
            & \multirow{2}{*}{Ref-LPIPS$\downarrow$} & Short-range & Long-range & \multirow{2}{*}{Ref-LPIPS$\downarrow$} & Short-range & Long-range\\
            & & consistency$\downarrow$ & consistency$\downarrow$ & & consistency$\downarrow$ & consistency$\downarrow$ \\
            \midrule
            ARF*~\cite{zhang2022arf} & 0.656 & \underline{0.027} & \underline{0.052} & 0.676 & \underline{0.032} & 0.222\\
            Ref-NPR*~\cite{zhang2023ref} & 0.645 & 0.028 & 0.064 & 0.669 & 0.034 & \underline{0.216}\\
            Texler et al.~\cite{texler2020interactive} & \textbf{0.593} & 0.035 & 0.072 & \textbf{0.621} & 0.045 & 0.243\\
            \rowcolor{LightGray}
            Ours & \underline{0.605} & \textbf{0.025} & \textbf{0.051} & \underline{0.639} & \textbf{0.030} & \textbf{0.211}\\
            \bottomrule
		\end{tabular}
    }
    \vspace{-2mm}
    \label{tab:dnerf-quantitative}
\end{table*}

\begin{figure}[t]
    \centering
    \includegraphics[width=0.47\textwidth]{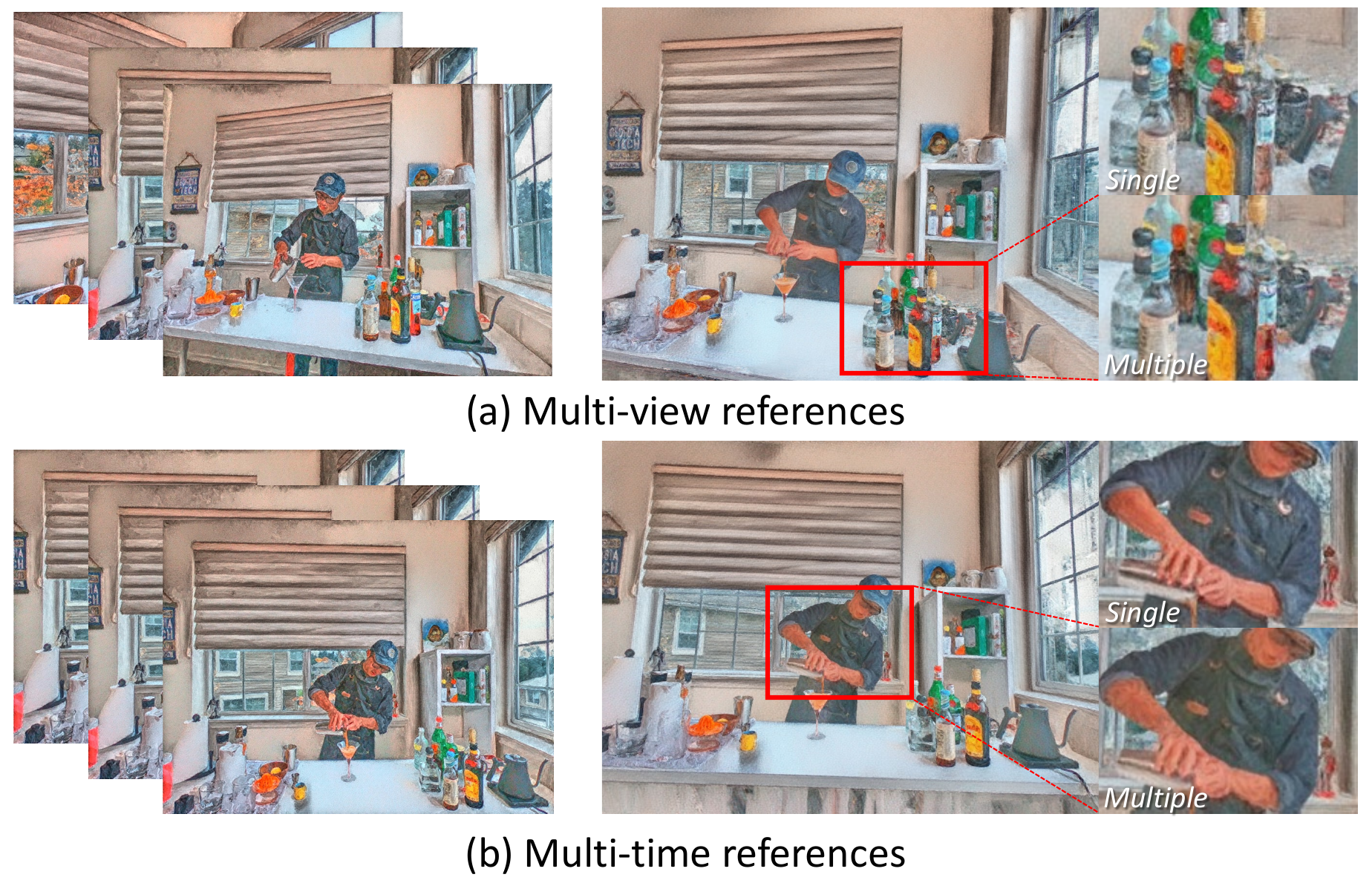}
    \caption{\textbf{Multi-references.} Our method is versatile and can also accept multiple references. We show that including additional references in spatial or temporal dimensions enriches the details and enhances the overall quality of the results.}
    % \vspace{-2em}
    \label{fig:multi-ref}
\end{figure}

\begin{figure*}[htbp]
    \centering
    \includegraphics[width=0.87\textwidth]{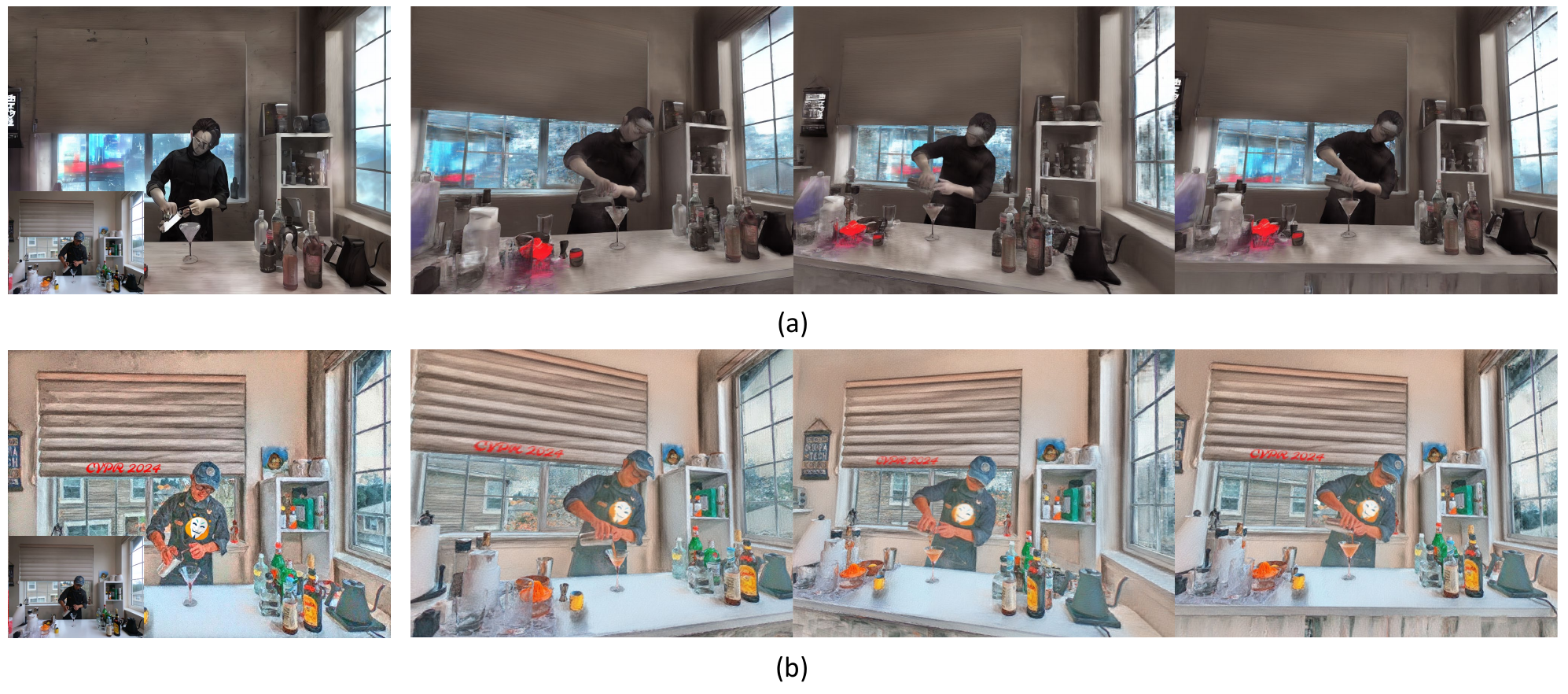}
    \caption{\textbf{Controllable stylization.} Our method inherently facilitates controllable stylization. (a) Besides neural style transfer methods such as NNST~\cite{kolkin2022neural}, we can leverage ControlNet~\cite{zhang2023adding} to generate or edit the reference image, and subsequently apply this modified reference to shape our dynamic 3D scenes. (b) Furthermore, our method enables localized edits to the reference image, making it possible to finetune or alter specific aspects of the scenes.}
    % \vspace{-2mm}
    \label{fig:controllable}
\end{figure*}

% \noindent\textbf{Datasets.}
% We assess the performance on two distinct datasets: the Plenoptic Video dataset~\cite{li2022neural} and the D-NeRF dataset~\cite{pumarola2021d}. The Plenoptic Video is derived from real-world scenarios, recorded with a multi-view camera system utilizing $21$ GoPro cameras. It captures video at a resolution of $2028 \times 2704$ and a frame rate of $30$ FPS. Each scene comprises $19$ synchronized $10$-second video clips, with $18$ designated for training and one for evaluation. On the other hand, the D-NeRF dataset features monocular videos with $360$-degree observations of synthetic objects.

\noindent\textbf{Experimental setup.}
We assess the performance on two distinct datasets: the Plenoptic Video dataset~\cite{li2022neural} and the D-NeRF dataset~\cite{pumarola2021d}.
To assess both the quality of stylization and the consistency between different novel views and novel timesteps of stylized dynamic 3D scenes, we employ two distinct settings. In one, we maintain a consistent time frame while generating novel views, and in the other, we maintain a consistent camera view while generating novel timesteps.
For the evaluation of reference-based stylization quality, we follow Wu et al.~\cite{wu2022ccpl} and adopt Ref-LPIPS. This metric quantifies the perceptual similarity between the style reference image and novel views or timesteps using LPIPS~\cite{zhang2018unreasonable}. To test the performance on cross-view consistency and temporal stability, we also adopt short-range consistency and long-range consistency used by Lai et al.~\cite{lai2018learning} and SNeRF~\cite{nguyen2022snerf}. Specifically, we warp a stylized view 
% $\mathcal{S}^{i}_{G_{\Theta}}$ 
to obtain a new view 
% $\mathcal{\hat{S}}^{i+\delta}_{G_{\Theta}}$ 
using the optical flow estimated by RAFT~\cite{teed2020raft}. We then compute the error between 
% $\mathcal{\hat{S}}^{i+\delta}_{G_{\Theta}}$ 
the warped view 
and the novel view or timestep 
% $\mathcal{S}^{i+\delta}_{G_{\Theta}}$ 
rendered from the stylized dynamic scene. To quantify short-range consistency, we compute the error between the $i^{th}$ and $(i + 1)^{th}$ video frames, while for long-range consistency, we compute the error between the $i^{th}$ and $(i + 7)^{th}$ frames.

\noindent\textbf{Quantitative comparisons.}
We show the quantitative results for the Plenoptic Video dataset and the D-NeRF dataset in Table~\ref{tab:nv3d-quantitative} and Table~\ref{tab:dnerf-quantitative}, respectively. The results indicate that our method achieves higher perceptual similarity in both real-world and synthetic scenarios. In addition, ARF*~\cite{zhang2022arf}, Ref-NPR*~\cite{zhang2023ref}, and our method demonstrate similar cross-view geometric consistency and temporal stability while Texler et al.~\cite{texler2020interactive} lags behind. 

\noindent\textbf{Qualitative comparisons.}
We present the visual comparisons in Fig.~\ref{fig:qualitative}. Given that ARF*~\cite{zhang2022arf} is an arbitrary style transfer approach, it introduces color mismatches between the reference image and the rendered view, leading to a perceptual gap, especially in comparison to reference-based stylization methods. While Ref-NPR*~\cite{zhang2023ref} can generate stylized novel views that preserve both geometric and semantic consistency with the given reference, it falls short in rendering novel times that achieve comparable visual quality to our method. The limitation arises from the fact that Ref-NPR does not account for the temporal propagation of style information from the stylized reference. Texler et al.~\cite{texler2020interactive} can generate high-quality stylized video sequences that retain structural details from the reference, yet it is plagued by flickering issues. In contrast, our method can output visually pleasing stylized novel views and times while maintaining semantic consistency with the provided reference across both spatial and temporal dimensions. 
% Readers are encouraged to view our supplementary video for better visual comparisons.

% \begin{figure*}[htbp]
%     \centering
%     \includegraphics[width=\textwidth]{figures/qualitative-nv3d.pdf}
%     \caption{\textbf{Qualitative comparisons on the Plenoptic Video dataset.} We compare our method with ARF*~\cite{zhang2022arf}, Ref-NPR*~\cite{zhang2023ref}, and Texler et al.~\cite{texler2020interactive}. In each case, the upper left image represents the reference view, generated from the photorealistic dynamic radiance field, while the lower left image depicts its corresponding stylized version. When compared to comparative alternatives, our method excels at generating visually pleasing stylized novel views and times while preserving semantic consistency with the provided reference image across both spatial and temporal dimensions.}
%     \vspace{-2mm}
%     \label{fig:qualitative-nv3d}
% \end{figure*}

% \begin{figure*}[htbp]
%     \centering
%     \includegraphics[width=\textwidth]{figures/qualitative-dnerf.pdf}
%     \caption{\textbf{Qualitative comparisons on the D-NeRF dataset.} In each case, the upper left image represents the reference view, generated from the photorealistic dynamic radiance field, while the lower left image depicts its corresponding stylized version. Here we showcase the stylized results of space-time view synthesis on dynamic 3D objects.}
%     \vspace{-2mm}
%     \label{fig:qualitative-dnerf}
% \end{figure*}

\begin{figure}[t]
    \centering
    \includegraphics[width=0.47\textwidth]{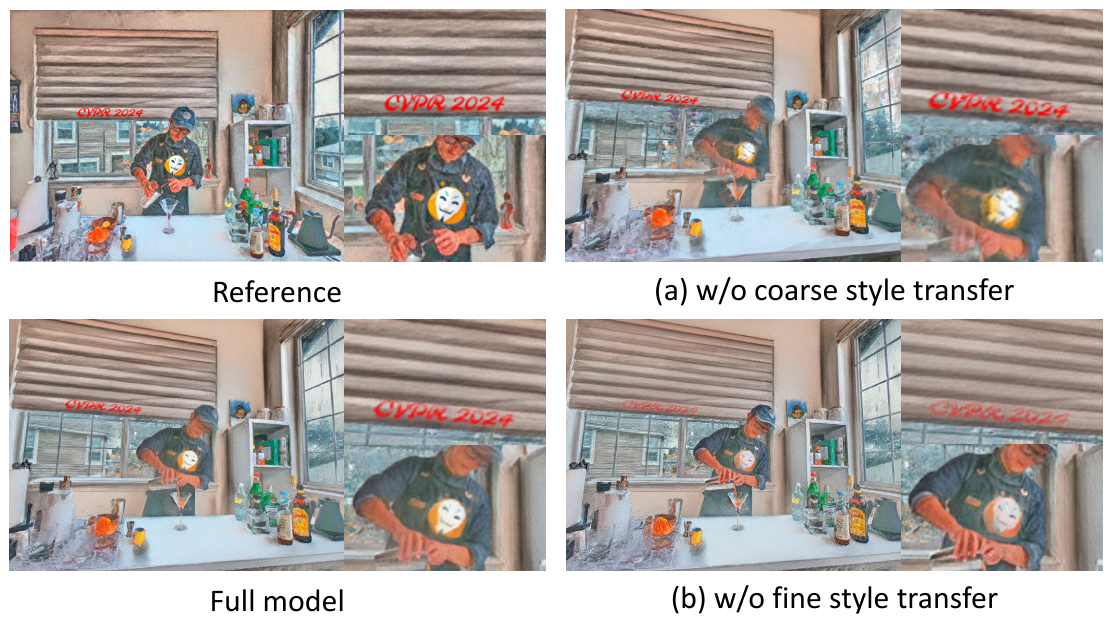}
    \caption{\textbf{Ablation study on each component of our method.}}
    % \vspace{-2mm}
    \label{fig:ablation}
\end{figure}

\noindent\textbf{Controllable stylization.}
In Fig.~\ref{fig:controllable}, we showcase the efficacy of our method in terms of controllability. Thanks to the reference-based nature, our method excels at producing diverse dynamic 3D scenes, aligning with the provided stylized reference. Besides neural style transfer methods such as NNST~\cite{kolkin2022neural}, we can also employ ControlNet~\cite{zhang2023adding} to produce or edit the reference image, and subsequently apply this modified reference to shape our dynamic 3D scenes. Furthermore, our method allows for localized edits to the reference image, enabling us to finetune or modify specific aspects of the scenes. This localized control empowers users to make precise adjustments to the visual elements, enhancing the overall customization and creative possibilities of our method.

% Besides neural style transfer methods such as NNST~\cite{kolkin2022neural}, we can also employ ControlNet~\cite{zhang2023adding} to produce or edit the reference image, and subsequently apply this modified reference to shape our dynamic 3D scenes. Furthermore, our method allows for localized edits to the reference image, enabling us to finetune or modify specific aspects of the scenes. This localized control empowers users to make precise adjustments to the visual elements, enhancing the overall customization and creative possibilities of our method. 

\noindent\textbf{Multi-references.}
In Sec.~\ref{sec:method}, our primary focus is on scenarios involving only a single reference image as input, simplifying the discussion for clarity. Nevertheless, it is worth noting that our method is versatile and can also accommodate multiple references. As illustrated in Fig.~\ref{fig:multi-ref}, we demonstrate how the inclusion of additional references in spatial or temporal dimensions enriches the details and enhances the overall quality of the results.

\subsection{User Study}
To further evaluate the performance of our method from a human perspective, we also conduct a user study comparing it with baseline methods, namely ARF*~\cite{zhang2022arf}, Ref-NPR*~\cite{zhang2023ref}, and Texler et al.~\cite{texler2020interactive}. Each participant is presented with a reference image, its stylized version, and two stylized videos: one generated by our method and the other from a randomly selected approach, with the order randomized. We invite a total of $56$ volunteers to select the method that exhibits superior perceptual quality, or none if it is hard to judge. The results are detailed in Table~\ref{tab:user-study}. These results clearly indicate a strong preference for our method.

\begin{table}[t]
    \footnotesize
    \centering 
    \caption{{\bf User study.} Pairwise comparison results indicate a strong preference for our method.}
    \resizebox{0.7\linewidth}{!}{
        \renewcommand\arraystretch{0.85}
		\begin{tabular}{lc}
            \toprule
            Comparison & Human preference \\
            \midrule
            ARF*~\cite{zhang2022arf} / Ours & 12.0\% / \textbf{88.0\%}  \\
            Ref-NPR*~\cite{zhang2023ref} / Ours & 22.6\% / \textbf{77.4\%} \\
            Texler et al.~\cite{texler2020interactive} / Ours & 30.0\% / \textbf{70.0\%} \\
            \bottomrule
		\end{tabular}
    }
    % \vspace{-2mm}
    \label{tab:user-study}
\end{table}

\subsection{Ablation Study}
To validate the effectiveness of each component, we conduct an ablation study on the Plenoptic Video dataset~\cite{li2022neural}. One can observe in Fig.~\ref{fig:ablation} (a) that the removal of the coarse style transfer module results in a noticeable decline in the quality of the stylized novel views and times. This decline occurs because the absence of the coarse style transfer module can render the training process vulnerable to getting stuck in local optima and impede its ability to provide meaningful stylization to occluded regions. In addition, as illustrated in Fig.~\ref{fig:ablation} (b), when the fine style transfer module is removed, the generated results tend to lack the preservation of style details. Together, these findings emphasize the contributions of both the coarse and fine style transfer modules to the overall quality and visual appeal of the results.

%% file: sec/6_conclusion.tex
\section{Conclusion}
\label{sec:conclusion}
In this paper, we have introduced a novel task of stylizing dynamic 3D scenes with one or a few stylized reference images as input. To this end, we present S-DyRF, a reference-based spatio-temporal stylization method for dynamic neural radiance fields. Our method effectively transfers the style information from the reference to the entire dynamic 3D scene, resulting in visually pleasing stylized novel views and times that maintain semantic consistency with the provided reference image across both spatial and temporal dimensions. To validate the effectiveness and superiority of our method, we conduct extensive experiments on both synthetic and real-world datasets. Moreover, thanks to the reference-based nature, our method empowers users to produce various stylized dynamic 3D scenes according to their preferences. We hope that S-DyRF can open up new avenues for 3D stylization and inspire further research in the domain of stylizing dynamic 3D scenes.